\documentclass[conference]{ieeeconf}

\usepackage{xcolor, soul}
\usepackage{float}
\sethlcolor{yellow}
\usepackage{graphicx} 
\usepackage{multirow}
\usepackage{booktabs}
\usepackage{bigstrut}
\usepackage{hyperref}
\usepackage{makecell}
\usepackage{amssymb}
\usepackage{amsmath}
\usepackage{tikz}
\graphicspath{ {./Figs/} } 
\begin{document}

%

\title{{A Tactile-enabled Grasping Method for Robotic Fruit Harvesting}}

\author {Hongyu Zhou$^{\dagger,1}$, Xing Wang$^{\dagger,1}$, Hanwen Kang$^{\dagger,1}$, Chao Chen$^{*,1}$  \\
$^{\dagger}$ These authors contribute equally\\
$^{1}$Department of Mechanical and Aerospace Engineering\\
Monash University\\
Clayton, VIC 3800, Australia \\
Email: chao.chen@monash.edu}

\maketitle 
\begin
{abstract}
In the robotic crop harvesting environment, foreign objects intrusion in the gripper workspace is frequently occurring and unignorable, however, rarely addressed.
This paper presents a novel intelligent robotic grasping method capable of handling obstacle interference, which is the first of its kind in the literature. The proposed method combines deep learning algorithms with low-cost tactile sensing hardware on a multi-DoF soft robotic gripper. Through experimental validations, the proposed method demonstrated promising performance in distinguishing various grasping scenarios. The 4-finger independently controlled gripper presented outstanding adaptability to handle various picking scenarios. The overall performance of this work indicated great potential for solving the robotic fruit harvesting challenges.
\end{abstract}
\IEEEpeerreviewmaketitle

\section{Introduction}
Robotic harvesting is one of the most challenging tasks in the agriculture industry. However, due to the complexity of the orchards environment, performing successful robotic harvesting is significantly difficult \cite{zhao2016review}. In the past two decades, significant efforts have been made in this field \cite{kang2020real, lin2021collision, tang2020recognition}, but few of them are proved to be efficient and reliable in in-field conditions. One major challenge of developing a successful fruit harvesting system is to enable robots with powerful capability in visual and tactile sensing, and therefore to reach and pick fruits like a human. As fruits on trees are most likely surrounded by stiff obstacles such as branches, trellis wires, or sprinkler lines. Those obstacles can not only reduce the success rate of the harvesting but also damage the robots and fruits. Recently, enormous efforts have been made in vision sensing, which enables robots to reach fruits by using visual sensors, such as cameras or ranged sensors \cite{kang2020fast, font2014proposal}. However, visual sensing can hardly work when the robotic arm reaches position which is to close to fruits. At this stage, tactile sensing is required to feedback that whether a robot can pick fruits successfully or not. Moreover, another issue of robotic harvesting is that the damage rate of fruits picked by robots is always exceeding an acceptable value, which leads to substantial loss of fruit yield and quality. Introducing tactile sensing in picking procedures is also promising to reduce the damage rate \cite{zhou2021intelligent}.

There are number of pilot researches conducted to apply tactile sensing with robotic grippers to provide self-sensing capacity \cite{jin2020triboelectric}. Elliott et al. \cite{donlon2018gelslim} developed a high-resolution tactile-sensing finger for grasping purposes. It is inspired by GelSight sensing and outputs the images with object information like shape and texture when contacting. Yancheng et al. \cite{wang2019flexible} proposed a flexible tactile sensor array with a 3 $\times$ 3 sensing unit and each has a five-electrode pattern's design. This tactile array can provide three-axis contact force perception while grasping different objects. Lingfeng et al. \cite{zhu2020development} designed a fully flexible tactile pressure sensor with the flexible graphene and silver composites as the sensing element and stretchable electrodes, respectively. The tactile sensor has relatively high sensitivity, wide sensing range, and considerable repeatability. The tactile sensor showed good performance after being integrated with robotic fingers to distinguish the cylinder and tennis ball during grasping. Linhan et al. \cite{yang2021learning} proposed a novel design of opto-electronic innervated tactile fingers to collect the tactile data such as normal force, torque, which were utilized to adjust the configuration of grasping. Besides, integrating vision and tactile sensing for the robotic grasping task also shows it advantages. For example, Di et al. \cite{guo2017robotic} combined visual and tactile sensing, while the former helps to detect the grasp rectangle from the visual images, and the latter helps to assess the stability of the grasping. 

Tactile sensing has been widely explored in many robotic grasping tasks, while only limited work focuses on robotic harvesting. To detach the target fruits from the plant, various grasping techniques with tactile sensing capacity have been implemented in the robotic harvesting application. Zhen et al. \cite{zhang2021hardness} proposed a two-finger gripper with a tactile sensor array with 4 $\times$ 6 elements installed inside of the fingers. The sensors can visualize the exertion force on the surface of the fruit, which can be further combined with displacement of the fingers to find the hardness of the target. The integration of tactile sensors for sensing hardness purposes helps the picking manipulation. Victoria et al. \cite{cortes2017integration} proposed novel non-destructive sensing for mango ripeness classification by combining the tactile sensors and near-infrared reflectance spectroscopy method. This ripeness detection helps to selectively pick the fruit during the robotic harvesting process.

From the aforementioned reviews, it is important to develop a gripper that can adjust its action based on tactile sensing feedback. In this research, we present a novel design of a fin-ray gripper with embedded tactile sensor and multi-DoF, which can adjust gripper state based on tactile feedback. To classify grasping status by using stress distribution from each finger, a robust perception algorithm and a deep-learning network is developed and validated. Lastly, we comprehensively evaluate our proposed system under controller experimental setup and in real harvesting action. The detailed contributions of this work are shown as follow,
\begin{itemize}
    \item We proposed a tactile enabled dexterous soft robotic gripper to address the obstacle interference challenge in the robotic fruit harvesting.
    \item We proposed a classification of the grasping pattern during robotic harvesting in orchard environment, which can be perceived by the tactile sensing array and processed by both traditional and deep-learning-based algorithms.  
    \item We validated the tactile sensing patterns and applied them to control the dexterous soft robotic gripper to achieve various grasping pattern under different grasping scenarios. 
  
\end{itemize}
 
For the rest of this paper, the design and fabrication process of our gripper is presented in Section II , and theoretical modelling and tactile sensing  is included in Section III. The experiments and results are presented in Section IV, followed by the conclusion in Section V. 

\section{Gripper design and fabrication}
\subsection{Mechanical Design}

The main objectives of the gripper design in the robotic fruit harvesting applications were: i) low fruit damage rate; ii) high harvest success rate; iii) ability of self-protection under semi-structured environment; iv) low cost. Low fruit damage rate demand gentle contact, while the high harvest success rate requires both essential grasping force and the capability to deal with obstacle interference. In terms of self-protection, the gripper has to be capable of sensing the external force during operation. 

\begin{figure}[h]
\centering
    \includegraphics[width=9cm,height=2cm]{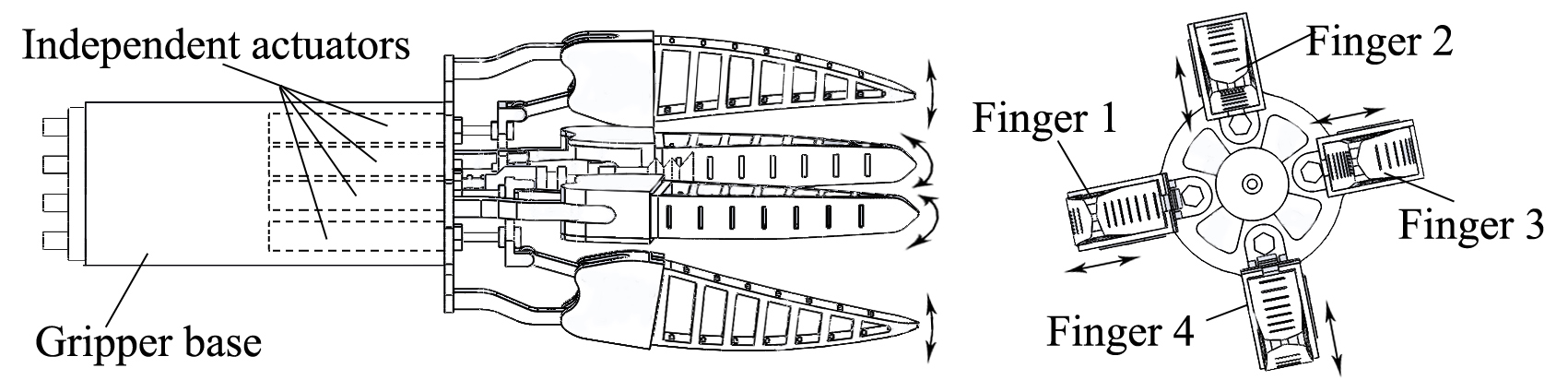}
    \caption{Mechanical design of the proposed multi-DoF soft gripper}
    \label{fig:mechanical design}
\end{figure}

With the above four objectives, a hybrid gripper actuated by four air cylinders with 4 flexible fingers is designed and prototyped, as shown in Figure \ref{fig:mechanical design}. Four actuators are introduced to provide essential grasping force while the flexible fingers are designed to achieve gentle contact. Specifically, the finger skeleton utilises the fin-ray effect for its outstanding shape adaptability, and the silicone skin cast on the surface of the finger can increase the friction for a stable grasp while distributing the pressure to mitigate the potential damage introduced by force concentration. Piezoresistive tactile sensing arrays (RX-M0404S) are integrated to enable the gripper with some extent of environment sensing ability.
To provide the adaptive gripper with self protection function as well as essential dexterity to deal with the obstacle interference, four fingers are designed to be able to act independently. Whenever an obstacle was detected by one finger, the central control system can react by either adjusting the approaching posture or releasing the finger and let the remaining fingers to complete the picking process.

\subsection{Finite Element Modeling}
\begin{figure}[ht]
    \centering
    \includegraphics[width=.48\textwidth]{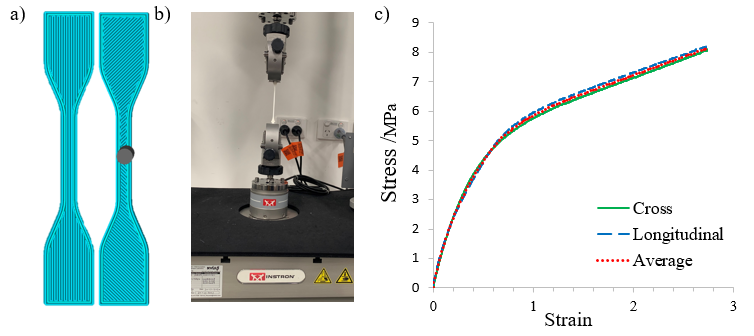}
    \caption{(a) Dumbbell test sample with longitudinal and cross pattern, (b) Instron tensile test machine, (c) Stress-strain curve of NinjaFlex}
    \label{fig:NinjaFlex}
\end{figure}

The grasping motion of the proposed soft gripper/finger can be simulated to predict the deformation of soft fingers \cite{zheng2019controllability, wang2020soft, wang2021bio, largilliere2015real} and provide the basic foundation for tactile sensing integration. To perform the simulation, the property of the hyperelastic material used (TPU: NinjaFlex) is initially tested by a uniaxial tensile test (ISO 37 standard). The dumbbell samples are printed and tested in two patterns, cross and longitudinal (Figure \ref{fig:NinjaFlex}a). The average stress and strain relation has been tested with Instron  Universal Tester E3000 (Figure \ref{fig:NinjaFlex}b and c). The average experimental data are fitted into three hyperelastic models, Yeoh, Ogden, and Mooney-Rivlin. Among these, the Ogden model has the best constitutive model with the parameter values $N_3$, $\mu_1$= 0.03829, $\alpha_1$ = 4.1352, $\mu_2$= 24.4601, $\alpha_2$ = 0.2123, $\mu_3$= 24.4613, $\alpha_3$ = 0.2122.

\begin{figure}[ht]
    \centering
    \includegraphics[width=.4\textwidth]{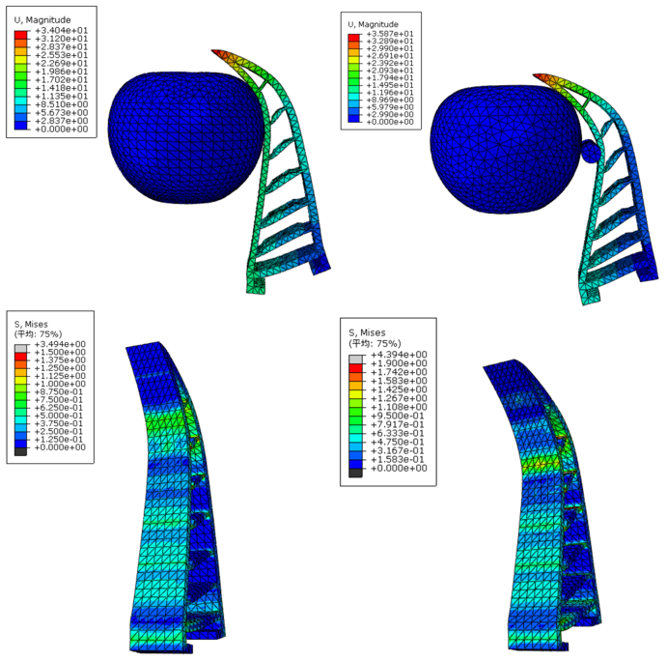}
    \caption{Simulation of soft fin ray finger's displacement when contacting (a) the apple model, (b) both apple and branch model, stress distribution of inner layer when contacting (c) apple model, (d) both apple and branch model}
    \label{fig:fea}
\end{figure}
The material is imported into the software Abaqus (Dassault System, MA). The soft finger is meshed using 4-node tetrahedron element. For the contact between the gripper and the target, the tangential contact behavior is model using the penalty method. Normal contact behavior is modeled as 'hard' contact. The lateral end of the soft finger is pinned and a displacement is provided on the medial side to simulate the actuation of fin-ray finger. Static analysis is performed with the nonlinear deformation enabled when grasping an apple model and apple model with branch interference, respectively. The final contracting state with the maximum displacement is shown in Figure \ref{fig:fea}a and b, where the soft finger contacts the target fruit or both the fruit and branch. The stress at the inner side where the tactile sensors can be attached is recorded as shown in Figure \ref{fig:fea}c and d. It needs to be noticed that once contacting the apple or both apple and branch, there is a significant stress concentration occurs in the bottom plane of the soft finger. The stress reaches the maximum where there is a contact area. At the position where the connection between the bottom plane and hinge connector occurs, the stress will increase as well. This simulation results verify the feasibility of integrating tactile sensors at the inner side of the soft finger to provide real-time feedback of the stress when contacting. These simulation also provides guidance on how and where the tactile sensors should be mounted to achieve better sensor reading. 

\subsection{Electronic Design}
\begin{figure}[h]
\centering
    \includegraphics[width=8.8cm,height=3cm]{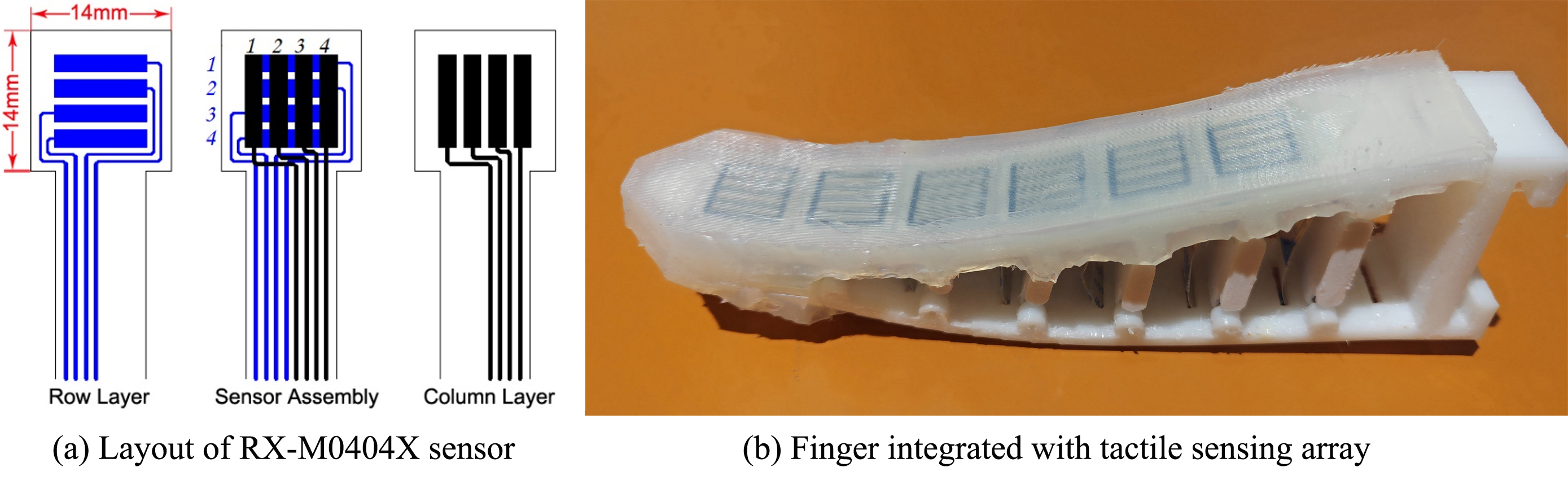}
     \caption{Layout of the tactile sensor and its integration of fin ray finger}
    \label{fig: tactile sensor}
\end{figure}

Piezoresistive effect-based sensors are applied due to their advantage in robustness and energy efficiency \cite{xu2018recent}. Specifically, 24 piezoresistive sensing arrays(RX-M0404S) are embedded on the gripper, with 6 sensors on each finger, as shown in a and b in Figure \ref{fig: tactile sensor}. Each sensor array has 4$\times$4 taxels in a 14mm $\times$14mm area. Each taxel can output resistance response to the external force. The detectable force range of taxel is from 0.2N to 20N.

\begin{figure}[h]
\centering
    \includegraphics[width=8.5cm,height=4cm]{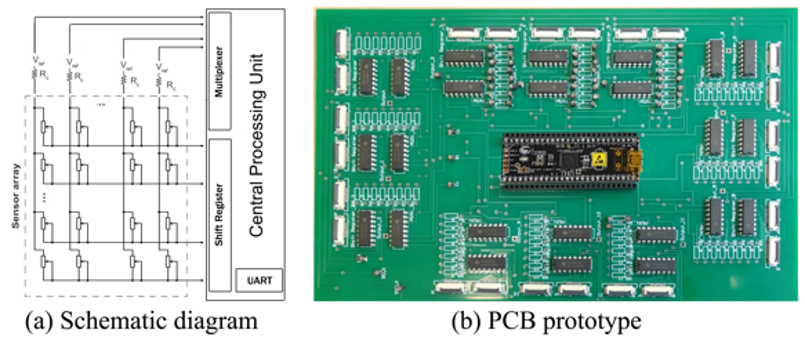} 
    \caption{Electrical circuit design of the data processing unit}
    \label{fig:Electrical design}
\end{figure}

To process data from the tactile sensors, a signal isolation circuit \cite{romano2011human} is used to overcome the potential extensive crosstalk between taxels. Then the measured resistor value will be uploaded to the Cypress PSoC controller via a selected channel of a multiplexer. The electrical schematic and the data processing circuit are shown in Figure \ref{fig:Electrical design}a and b, respectively.

\section{Sensing Algorithm}

\begin{figure}[h]
\centering
\includegraphics[width=.3\textwidth]{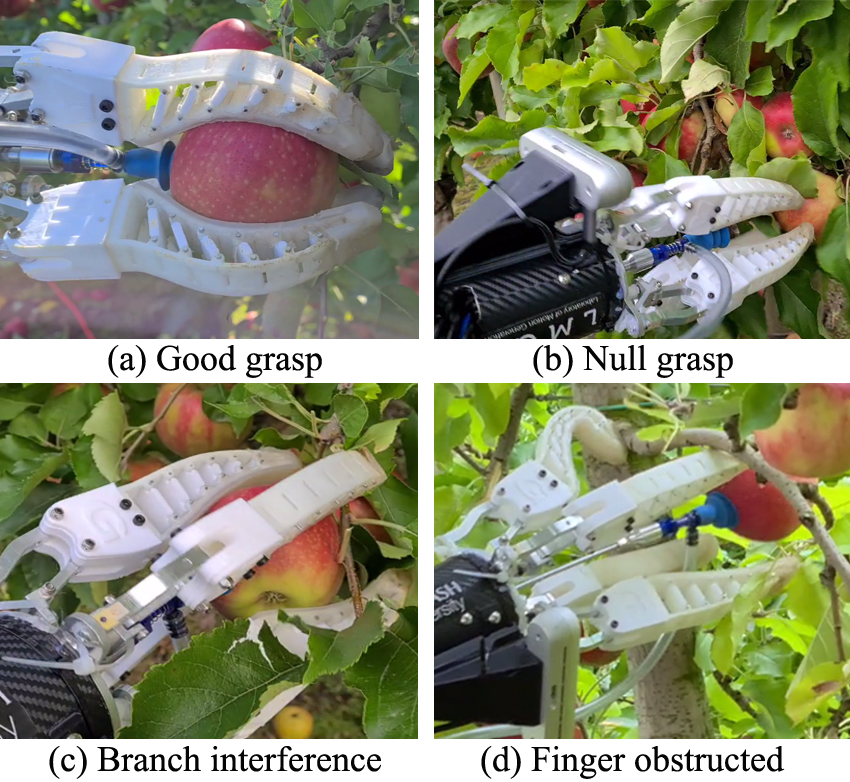} 
    \caption{Four grasping patterns defined during robotic harvesting}
    \label{fig:scenarios}
\end{figure}

We consider four grasping statuses in this work: null grasp, good grasp, finger interference grasp, and obstacle obstructed grasp, as shown in Figure \ref{fig:scenarios}. Specifically, null grasp means no fruits are retrieved in the gripper, good grasp means all fingers have stably held the fruits, finger interference grasp means a branch gets caught in between one or more fingers of the gripper, blocked grasp indicates that one or more fingers are obstructed by branches or other obstacles such as trellis wires, trellis support beam, etc. 

\subsection{State-estimation algorithm}

To distinguish the are status of the gripper during the fruit picking process, three assumptions are established. First of all, the pressure value detected by the tactile sensing arrays of the four fingers against a null grasp are assumed to be much smaller than which is in a good grasp. Secondly, in the obstacle grasp, one or two fingers will be stopped by the obstacle when approaching the target fruit, which means the obstructed finger or fingers will output significantly earlier change than the other fingers. Last but most important, for the branch interference category, according to the theoretical modelling analysis, a branch being grasped in between a finger and the target fruit will trigger force concentration near the contact area. We assume such force concentration would lead to two outcomes: on the one hand, the pressure value in the force concentration area tend to be higher than an adjacent area; on the other hand, the change of the pressure value in a certain time frame would be much sharper. 
With these assumptions, moving variance of the pressure value on each taxel was taken as an indicator to characterize the rate of pressure change. 

\begin{figure}[h]
\centering
\includegraphics[width=6cm,height=3cm]{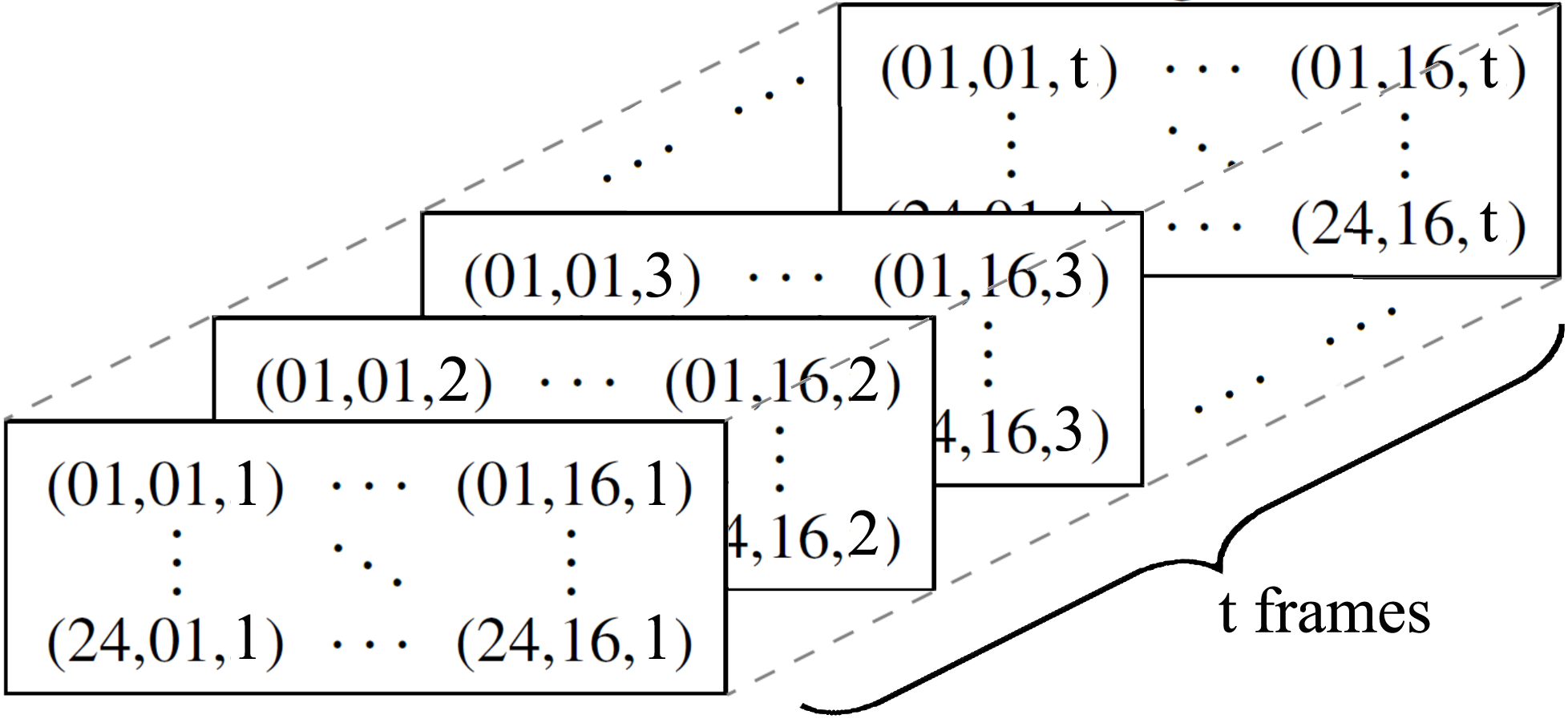} 
    \caption{Time series datasets}
    \label{fig:datasets}
\end{figure}

To analyse the moving variance of the pressure value during the grasping process, the output of the 384 taxel feedback was fed into a 24$\times$16$\times$t matrix, where t is the number of time frames, as shown in Figure \ref{fig:datasets}. The interval of each time frame was set as 60ms. To monitor the changes of the values of each matrix element, the 24$\times$16$\times$t matrix is reshaped to a 384$\times$t matrix (M$_{1}$) after data normalization. A moving average of every four data points was then applied to the normalized data to filter out the noise, after which a moving variance was calculated against each matrix element, four maximum moving variance on the taxels of four fingers was then selected for comparison.

\begin{equation}
M_1=\begin{bmatrix}
\sigma_{1,1}^2 & \cdots  & \sigma_{1,t}^2  \\
\vdots              & \ddots  & \vdots  \\
\sigma_{384,1}^2 & \cdots  & \sigma_{384,t}^2 \\
\end{bmatrix}
\end{equation}

where, 
\begin{equation}
\sigma_i^2=\frac{1}{n}{\sum_{k=1}^n(v_j-\bar{v})^2}
\end{equation}

Various conventional statistic analysis methods were tested, including moving average, moving variance, Fast Fourier Transform, and power spectral density.

\subsection{State-estimation Network}
\begin{figure}[h]
\centering
\includegraphics[width=8.5cm,height=3.8cm]{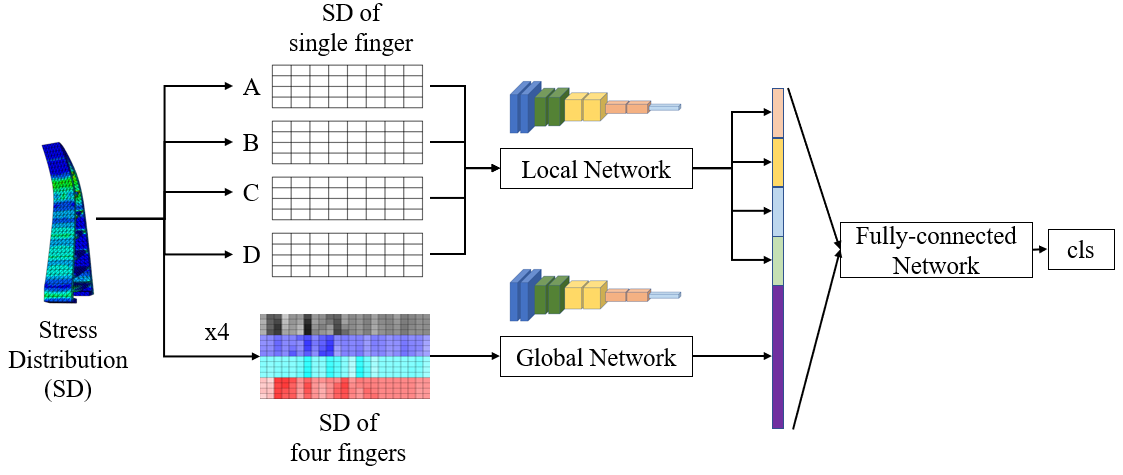} 
\caption{Deep-touch CNN network for grasping pattern classification}
\label{fig:NN}
\end{figure}

The embedded tactile sensors in fin-ray fingers can continuously feedback a stress distribution matrix in $24 \times 16$ pixels. To simplify the signal preprocessing from conventional analysis, a CNN model Deep-touch is also designed to predict the grasping status. Deep-touch includes three networks, a local finger network to predict the status of a single finger, a global network to extract features of global stress distributions of four fingers, and a fully-connected network to combine features from local and global networks to predict current grasping status (Figure \ref{fig:NN}).
Local finger network applies a ResNet-18 model, while the pooling layer kernel is changed to 2 $\times$ 1. Each local finger network generates a 1 $\times$ 64 vector. The global network also uses ResNet-18 model while the input is the stress distribution of four fingers and the output is a 1 $\times$ 256 feature vector. The feature vectors from four fingers and the global network are concatenated together, generating a feature vector of 1 $\times$ 512. This vector is then fed into the fully-connected network to predict the current status of the gripper.

\section{Experiment and Results}
 
\subsection{Experiment on tactile-enabled grasping}

\subsubsection{Experiment Setup}

To validate the proposed method, the experiment is set up: the sensor-integrated gripper was fixed on a desk, then grasp tests were conducted on four scenarios. Twelve apples of different varieties were used as target objects to be grasped by the gripper. Each apple was grasped multiple times in different orientations. To make the branch settings close to reality, the branches were also set to different orientations and positions. 

    \begin{figure}[h]
    \includegraphics[width=8.5cm,height=4.5cm]{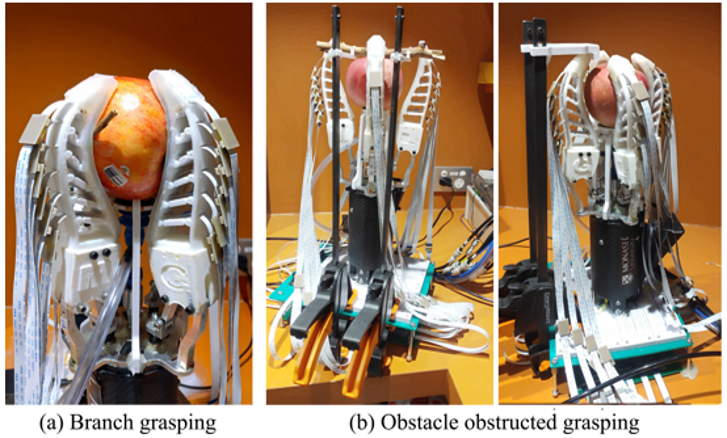}
    \caption{Lab experiment setup with (a) branch grasping, (b)obstacle obstructed grasping}
    \label{fig:Experiment setup}
    \end{figure}

\subsubsection{Experiment on conventional methods}
200 grasp tests were implemented to validate the proposed conventional method, as shown in Figure \ref{fig:Experiment setup}. Specifically, 96 grasps for branch interference, 48 grasps for good grasp, 30 for null grasp and 26 for finger obstructed grasp.

For all the tests, the proposed moving variance method achieved an overall accuracy of 100 percent in detecting the 30 null grasps, 92 percent in detecting the 26 obstacle grasps, and 75 percent in determining which finger has a branch grasped among the 96 branch interference grasps. The detail of the test result is shown in Table \ref{tab:1}.

\begin{figure}[h]
\includegraphics[width=8.8cm,height=7cm]{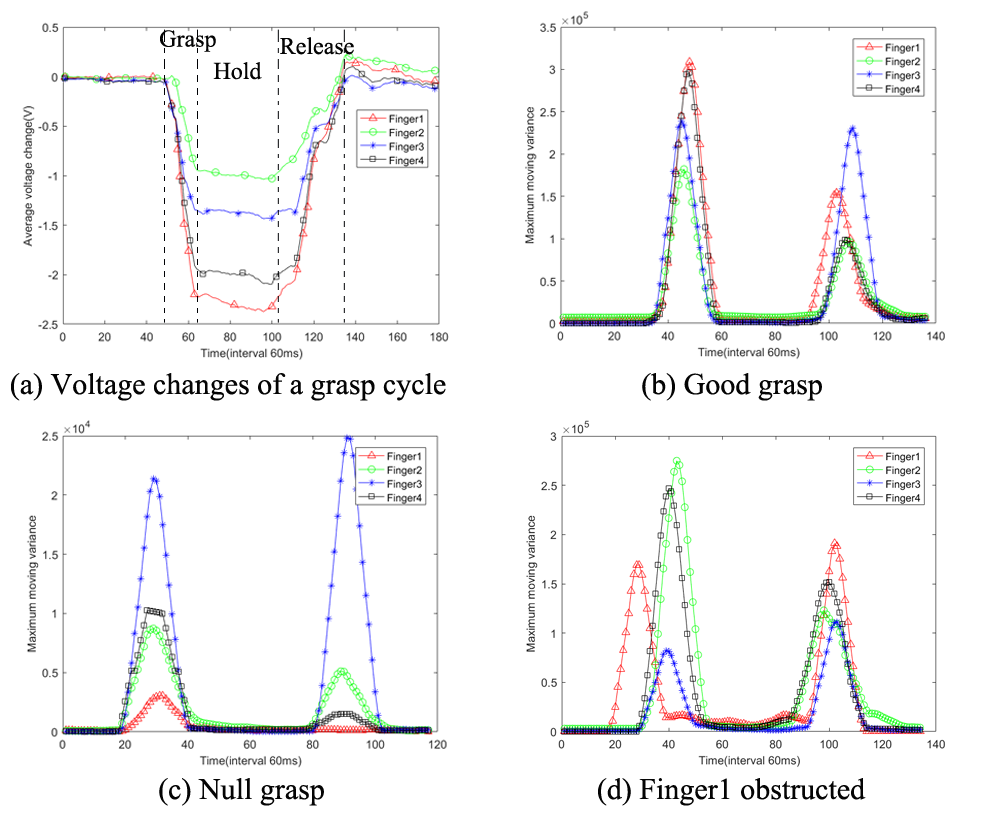}
    \caption{Voltage and variance change during grasping process among different scenarios}
    \label{fig:grasp cycle and data analysis} 
\end{figure}

Figure \ref{fig:grasp cycle and data analysis} shows the moving variance changes over the entire grasp process (approaching-grasp-hold-release), the amplitude of the maximum moving variance calculated on the four fingers of the null grasp are significantly smaller than other scenarios. While in the obstacle grasp scenario, the variance change on four fingers shows apparent asynchronicity, representing that the deformation of one finger occurs remarkably earlier than other fingers.

\begin{figure}[h]
\includegraphics[width=8.8cm,height=7cm]{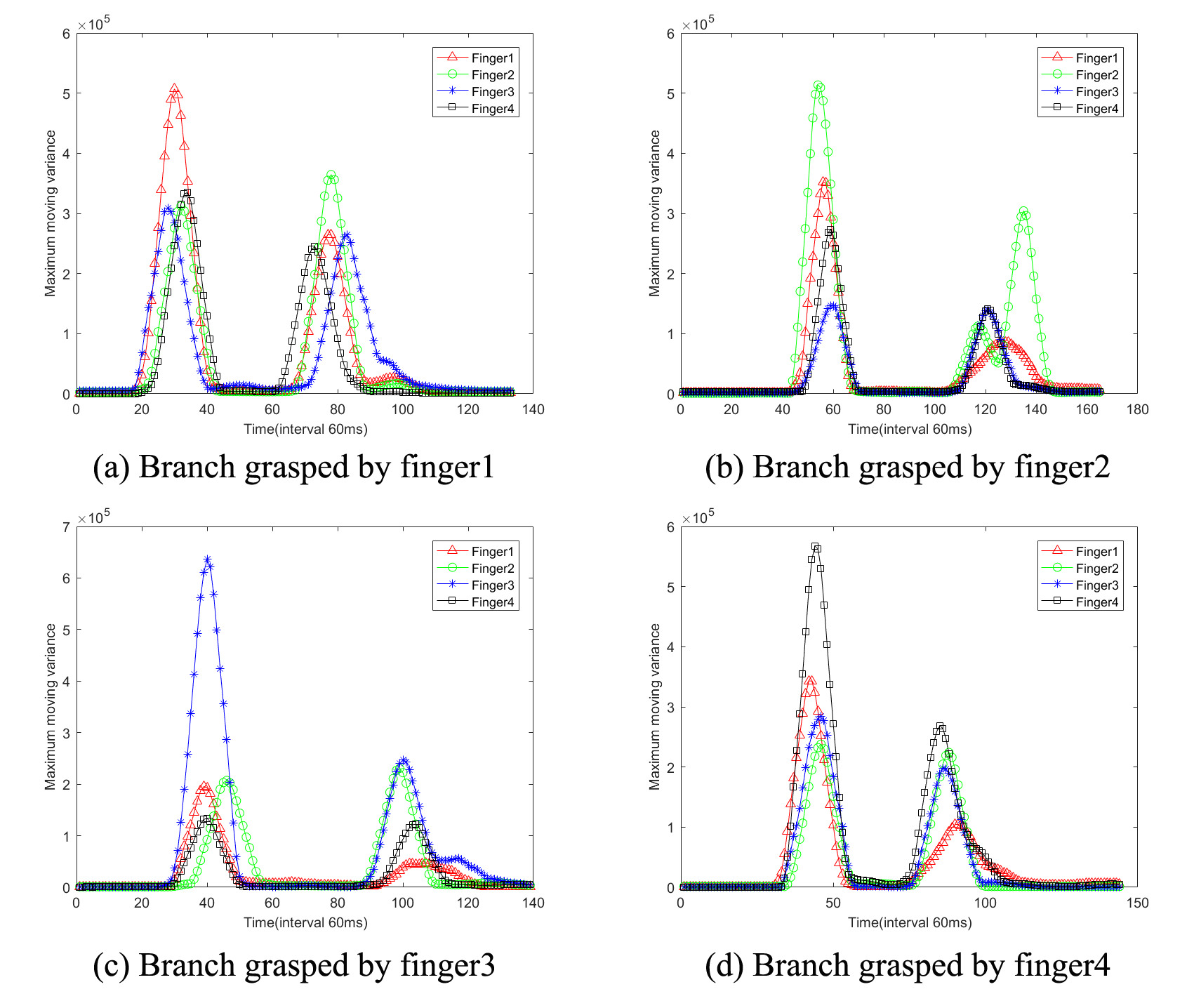}
    \caption{Variance change when one finger grasped a branch}
    \label{fig:branch interference}
\end{figure}

 The moving variance changes of finger interference grasp where a branch grasped by different fingers is presented in Figure \ref{fig:branch interference} a, b, c, and d, respectively, from which we can see that the finger with a branch been grasped in output a much higher moving variance in the grasp phase.

In terms of differentiating good grasp from branch interference grasp, various conventional statistic analysis methods were tried, including moving average, moving variance, Fast Fourier Transform, power spectral density, each of these conventional feature only works for a certain condition but not able to cover all. This indicates that some hidden features were yet to be fully extracted to distinguish the difference between good grasp and branch interference grasp. Therefore the deep learning methods are required to bridge the gap.

\subsubsection{Experiment on Deep-learning methods}
To evaluate the performance of the trained deep-touch classification network, the same data set is utilized as compared with conventional method. The stress distribution image of each of the four fingers and the merged results are input into the classification network. The predicted labels are output and compared with the ground truth label to verify the accuracy of deep-touch network. The classification accuracy for each grasping status is summarised in Table \ref{tab:1} as well. It can be seen that the deep-touch network achieves the same classification accuracy on null grasp status, as compared with the convention algorithm. This is due to the factor that the features are quite obvious to extract for both convention and neural networks. The null grasp shows a much smaller voltage change, thus a much less dense color visualization in the stress distribution image. As for the other grasp status, the deep-learning-based algorithm shows its superior performance in  classification accuracy. This is due to the fact that our network has both global and local networks which help with abundant feature extractions. The deep-touch network demonstrated a minimum 3.8\% accuracy improvement while identifying these three statuses. The average accuracy improvement reaches to 15.13\% among these three. The deep-learning-based network shows great superiority in differentiating good grasp and branch interference grasp. An overall 89.4\% accuracy in distinguishing the four scenarios is quite promising.
\begin{table}[htbp]
  \centering
  \caption{Experiment result of the proposed conventional method}
    \begin{tabular}{p{10mm}llccc}
    \toprule
    \multicolumn{1}{p{4.8em}}{\multirow{1}[4]{*}{\textbf{Scenarios}}} & \multicolumn{1}{c}{\multirow{1}[4]{*}{\textbf{Category}}} & \multicolumn{1}{c}{\multirow{2}[4]{*}{\textbf{\shortstack{Number\\of\\grasps}}}} & \multicolumn{2}{c}{\textbf{Detection accuracy}} \\
\cmidrule{4-5}    &    &    & \multicolumn{1}{p{6em}}{\textbf{\shortstack{Conventional\\method}}} & \multicolumn{1}{p{4em}}{\textbf{\shortstack{Deep\\learning}}} \\
    \midrule
    \multirow{2}[4]{*}{Null grasp} & \shortstack{Without\\ leaves} & 15 & \multirow{2}[4]{*}{96.6\%} & \multirow{2}[4]{*}{96.6\%} \\
\cmidrule{2-3}       & \shortstack{With\\ leaves} & 15 &    &  \\
    \midrule
    \multicolumn{1}{p{5em}}{\shortstack{Finger\\obstructed\\ grasp}} & N/A & 26 & 88.5\% & 92.3\% \\
    \midrule
    Good grasp & N/A & 48 & 52.1\% & 85.4\% \\
    \midrule
    \multicolumn{1}{l}{\multirow{4}[8]{*}{\shortstack{Branch\\interference\\grasp}}} & Finger 1 & 24 & \multirow{4}[8]{*}{75.0\%} & \multirow{4}[8]{*}{83.3\%} \\
\cmidrule{2-3}       & Finger 2 & 24 &    &  \\
\cmidrule{2-3}       & Finger 3 & 24 &    &  \\
\cmidrule{2-3}       & Finger 4 & 24 &    &  \\
    \bottomrule
    \end{tabular}%
  \label{tab:1}%
\end{table}%

\subsection{Experiment on Harvesting system}
\begin{figure}[h]
    \includegraphics[width=0.4\textwidth]{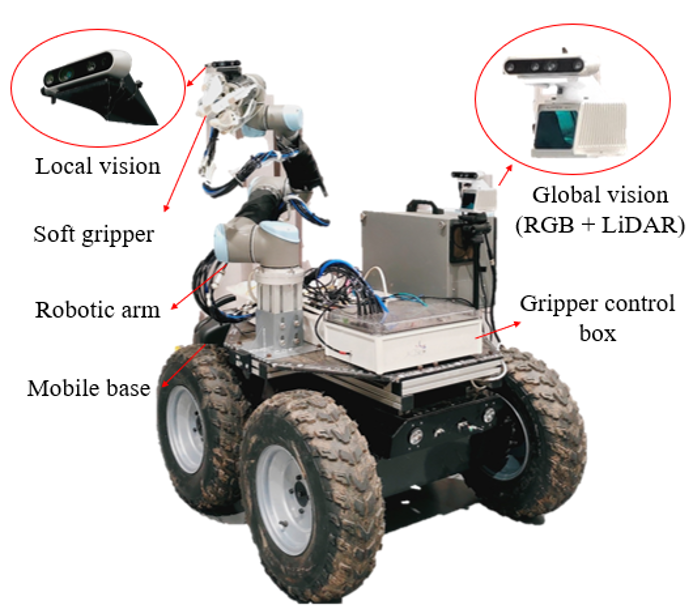}
    \caption{Monash robotic fruit retrieving system}
    \label{fig:system}
    \end{figure}

\begin{figure}[ht]
    \centering
    \includegraphics[width=.4\textwidth]{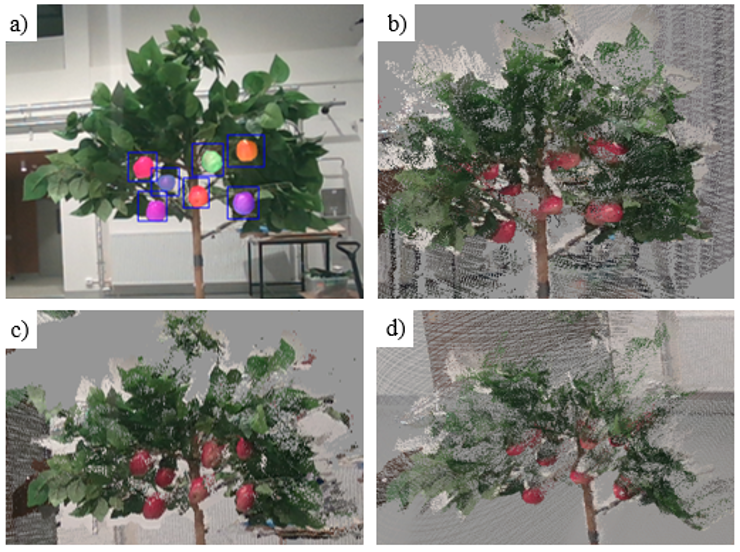}
    \caption{(a) RGB image captured by the global camera with apple detection and segmentation, colored point clouds of apple tree model from RGB-LiDAR fusion (b) left view, (c) centre view, (d) right view }
    \label{fig:fusion}
\end{figure}

A robotic system is used to demonstrate tactile-enabled grasping in the lab environment. The system includes four subsystems, a vehicle, a 6-DoF universal robotic arm, the proposed gripper, and a vision system, as shown in Figure \ref{fig:system}. Our robot has two vision blocks: global and a local block. The global vision includes a DJI-Livox Mid-70 LiDAR and an RGB camera. The color image is utilized to perform the 2D fruit recognition and segmentation (Figure \ref{fig:fusion} a), which is also calibrated and fused with point cloud from the LiDAR, as shown in Figure \ref{fig:fusion} b, c, and d.  Local vision has a depth camera on the end-effector to perform second-time processing. A geometry-aware detection network is applied in both blocks, which can localise and predict the reach angle of each apple. Octomap is used to modelling the occupancy space by using point cloud from the vision. MoveIt! framework is used in arm planning and execution. 

\begin{figure}[h]
    \includegraphics[width=0.45\textwidth]{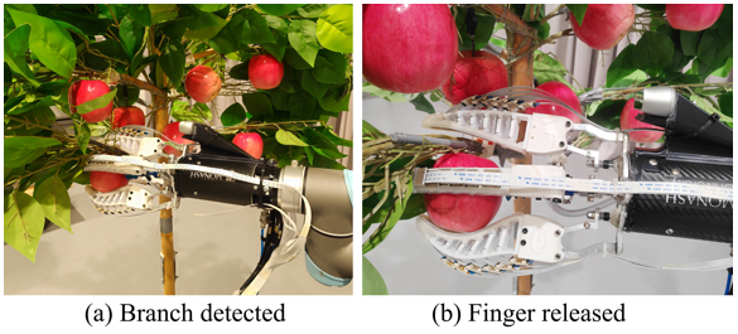}
    \caption{Adjusted grasping pattern of the multi-DoF gripper}
    \label{fig: demo}
    \end{figure}

The experiments were conducted in the lab environment, and seven apples were hanging on the fake apple tree to simulate the different occlusion. The targets were detected and validated by the global and local visual system with position information, the proposed multi-DoF gripper was then manipulated to approach it. The gripper was actuated to grasp when reaching the front position of the target fruit, in this case, the tactile sensors received and sent the contact force information. The deep-learning-based classification network classified the grasping pattern into one of the four patterns. The multi-DoF gripper can adjust the modes depending on different patterns classified by the network. For example, it opens and re-attempts to grasping once there is a null/obstacle grasp pattern. It detaches the grasped fruit once the good grasp is received. In the case branch interference grasp is detected (shown in Figure \ref{fig: demo}a), the traditional algorithm was utilized afterward to locate the finger where there is a branch interference. After the finger was identified, that finger was pneumatically controlled to open and release the grasped branch, as shown in Figure \ref{fig: demo}b. The detach continued to remove the target fruit off the tree.

\section{Conclusion}
This work developed an intelligent robotic grasping method based on sensing algorithms and a novel designed soft gripper prototype, which can distinguish various grasping scenarios during the robotic harvesting process and adjust its grasping action based on the multi-DoF mechanism.
The proposed sensing-based grasping method is the first of its kind in the literature to handle the branch interference challenge. Such method can be further applied to broader fields, whenever there are foreign objects intrude into the gripper workspace.
Besides, the deep learning algorithms have been developed to achieve accurate grasping status detection, regardless of the noises generated by the silicone skin of the gripper fingers. Compared with the conventional method, the proposed method presented an average 15.13\% increase in the classification accuracy.
A robotic harvesting system has been built to included different core components. The gripper is able to approach and contact the target fruit based on the robust vision and manipulation system. The demonstration of the proposed tactile-enabled grasping method has been validated trough experiments. The grasp patterns can be identified and utilized to control the soft gripper to adjust the actions.


\section*{Acknowledgment}

We gratefully acknowledge the financial support from Australian Research Council (ARC ITRH IH150100006).
We would like to thank Mr. Cooper Gerwing, Mr. Charles Troeung, Dr. Wesley Au, Dr. Shao Liu and Dr. Godfrey Keung in the Laboratory of Motion Generation Analysis at Monash University for their assistance on this work.


\bibliographystyle{IEEEtran}
\bibliography{ref}

\begin{thebibliography}{10}
\providecommand{\url}[1]{#1}
\csname url@samestyle\endcsname
\providecommand{\newblock}{\relax}
\providecommand{\bibinfo}[2]{#2}
\providecommand{\BIBentrySTDinterwordspacing}{\spaceskip=0pt\relax}
\providecommand{\BIBentryALTinterwordstretchfactor}{4}
\providecommand{\BIBentryALTinterwordspacing}{\spaceskip=\fontdimen2\font plus
\BIBentryALTinterwordstretchfactor\fontdimen3\font minus
  \fontdimen4\font\relax}
\providecommand{\BIBforeignlanguage}[2]{{%
\expandafter\ifx\csname l@#1\endcsname\relax
\typeout{** WARNING: IEEEtran.bst: No hyphenation pattern has been}%
\typeout{** loaded for the language `#1'. Using the pattern for}%
\typeout{** the default language instead.}%
\else
\language=\csname l@#1\endcsname
\fi
#2}}
\providecommand{\BIBdecl}{\relax}
\BIBdecl

\bibitem{zhao2016review}
Y.~Zhao, L.~Gong, Y.~Huang, and C.~Liu, ``A review of key techniques of
  vision-based control for harvesting robot,'' \emph{Computers and Electronics
  in Agriculture}, vol. 127, pp. 311--323, 2016.

\bibitem{kang2020real}
H.~Kang, H.~Zhou, X.~Wang, and C.~Chen, ``Real-time fruit recognition and
  grasping estimation for robotic apple harvesting,'' \emph{Sensors}, vol.~20,
  no.~19, p. 5670, 2020.

\bibitem{lin2021collision}
G.~Lin, L.~Zhu, J.~Li, X.~Zou, and Y.~Tang, ``Collision-free path planning for
  a guava-harvesting robot based on recurrent deep reinforcement learning,''
  \emph{Computers and Electronics in Agriculture}, vol. 188, p. 106350, 2021.

\bibitem{tang2020recognition}
Y.~Tang, M.~Chen, C.~Wang, L.~Luo, J.~Li, G.~Lian, and X.~Zou, ``Recognition
  and localization methods for vision-based fruit picking robots: A review,''
  \emph{Frontiers in Plant Science}, vol.~11, p. 510, 2020.

\bibitem{kang2020fast}
H.~Kang and C.~Chen, ``Fast implementation of real-time fruit detection in
  apple orchards using deep learning,'' \emph{Computers and Electronics in
  Agriculture}, vol. 168, p. 105108, 2020.

\bibitem{font2014proposal}
D.~Font, T.~Pallej{\`a}, M.~Tresanchez, D.~Runcan, J.~Moreno, D.~Mart{\'\i}nez,
  M.~Teixid{\'o}, and J.~Palac{\'\i}n, ``A proposal for automatic fruit
  harvesting by combining a low cost stereovision camera and a robotic arm,''
  \emph{Sensors}, vol.~14, no.~7, pp. 11\,557--11\,579, 2014.

\bibitem{zhou2021intelligent}
H.~Zhou, X.~Wang, W.~Au, H.~Kang, and C.~Chen, ``Intelligent robots for fruit
  harvesting: Recent developments and future challenges,'' 2021.

\bibitem{jin2020triboelectric}
T.~Jin, Z.~Sun, L.~Li, Q.~Zhang, M.~Zhu, Z.~Zhang, G.~Yuan, T.~Chen, Y.~Tian,
  X.~Hou \emph{et~al.}, ``Triboelectric nanogenerator sensors for soft robotics
  aiming at digital twin applications,'' \emph{Nature communications}, vol.~11,
  no.~1, pp. 1--12, 2020.

\bibitem{donlon2018gelslim}
E.~Donlon, S.~Dong, M.~Liu, J.~Li, E.~Adelson, and A.~Rodriguez, ``Gelslim: A
  high-resolution, compact, robust, and calibrated tactile-sensing finger,'' in
  \emph{2018 IEEE/RSJ International Conference on Intelligent Robots and
  Systems (IROS)}.\hskip 1em plus 0.5em minus 0.4em\relax IEEE, 2018, pp.
  1927--1934.

\bibitem{wang2019flexible}
Y.~Wang, J.~Chen, and D.~Mei, ``Flexible tactile sensor array for slippage and
  grooved surface recognition in sliding movement,'' \emph{Micromachines},
  vol.~10, no.~9, p. 579, 2019.

\bibitem{zhu2020development}
L.~Zhu, Y.~Wang, D.~Mei, and C.~Jiang, ``Development of fully flexible tactile
  pressure sensor with bilayer interlaced bumps for robotic grasping
  applications,'' \emph{Micromachines}, vol.~11, no.~8, p. 770, 2020.

\bibitem{yang2021learning}
L.~Yang, X.~Han, W.~Guo, F.~Wan, J.~Pan, and C.~Song, ``Learning-based
  optoelectronically innervated tactile finger for rigid-soft interactive
  grasping,'' \emph{IEEE Robotics and Automation Letters}, vol.~6, no.~2, pp.
  3817--3824, 2021.

\bibitem{guo2017robotic}
D.~Guo, F.~Sun, B.~Fang, C.~Yang, and N.~Xi, ``Robotic grasping using visual
  and tactile sensing,'' \emph{Information Sciences}, vol. 417, pp. 274--286,
  2017.

\bibitem{zhang2021hardness}
Z.~Zhang, J.~Zhou, Z.~Yan, K.~Wang, J.~Mao, and Z.~Jiang, ``Hardness
  recognition of fruits and vegetables based on tactile array information of
  manipulator,'' \emph{Computers and Electronics in Agriculture}, vol. 181, p.
  105959, 2021.

\bibitem{cortes2017integration}
V.~Cort{\'e}s, C.~Blanes, J.~Blasco, C.~Ortiz, N.~Aleixos, M.~Mellado,
  S.~Cubero, and P.~Talens, ``Integration of simultaneous tactile sensing and
  visible and near-infrared reflectance spectroscopy in a robot gripper for
  mango quality assessment,'' \emph{Biosystems Engineering}, vol. 162, pp.
  112--123, 2017.

\bibitem{zheng2019controllability}
G.~Zheng, O.~Goury, M.~Thieffry, A.~Kruszewski, and C.~Duriez,
  ``Controllability pre-verification of silicone soft robots based on
  finite-element method,'' in \emph{2019 International Conference on Robotics
  and Automation (ICRA)}.\hskip 1em plus 0.5em minus 0.4em\relax IEEE, 2019,
  pp. 7395--7400.

\bibitem{wang2020soft}
X.~Wang, A.~Khara, and C.~Chen, ``A soft pneumatic bistable reinforced actuator
  bioinspired by venus flytrap with enhanced grasping capability,''
  \emph{Bioinspiration \& Biomimetics}, vol.~15, no.~5, p. 056017, 2020.

\bibitem{wang2021bio}
X.~Wang, H.~Zhou, H.~Kang, W.~Au, and C.~Chen, ``Bio-inspired soft bistable
  actuator with dual actuations,'' \emph{Smart Materials and Structures}, 2021.

\bibitem{largilliere2015real}
F.~Largilliere, V.~Verona, E.~Coevoet, M.~Sanz-Lopez, J.~Dequidt, and
  C.~Duriez, ``Real-time control of soft-robots using asynchronous finite
  element modeling,'' in \emph{2015 IEEE International Conference on Robotics
  and Automation (ICRA)}.\hskip 1em plus 0.5em minus 0.4em\relax IEEE, 2015,
  pp. 2550--2555.

\bibitem{xu2018recent}
F.~Xu, X.~Li, Y.~Shi, L.~Li, W.~Wang, L.~He, and R.~Liu, ``Recent developments
  for flexible pressure sensors: A review,'' \emph{Micromachines}, vol.~9,
  no.~11, p. 580, 2018.

\bibitem{romano2011human}
J.~M. Romano, K.~Hsiao, G.~Niemeyer, S.~Chitta, and K.~J. Kuchenbecker,
  ``Human-inspired robotic grasp control with tactile sensing,'' \emph{IEEE
  Transactions on Robotics}, vol.~27, no.~6, pp. 1067--1079, 2011.

\end{thebibliography}

%




\end{document}